# Hybrid Deep Learning for Legal Text Analysis: Predicting Punishment Durations in Indonesian Court Rulings

Muhammad A. Ibrahim* [1], Alif T. Handoyo[1], and Maria S. Anggreainy[2,]

[1] Department of Computer Science, School of Computer Science, Bina Nusantara University, Jakarta 11540, Indonesia
[2] Computer Science Department, Doctor of Computer Science, Bina Nusantara University, Jakarta 11540, Indonesia
Email: muhammad.amien@binus.ac.id (M.A.I.); alif.handoyo@binus.ac.id (A.T.H.); maria.susan001@binus.ac.id (M.S.A)
*Corresponding author

***Abstract*— Limited public understanding of legal processes and inconsistent verdicts in the Indonesian court system led to widespread dissatisfaction and increased stress on judges. This study addresses these issues by developing a deep learning-based predictive system for court sentence lengths. Our hybrid model, combining CNN and BiLSTM with attention mechanism, achieved an R-squared score of 0.5893, effectively capturing both local patterns and long-term dependencies in legal texts. While document summarization proved ineffective, using only the top 30% most frequent tokens increased prediction performance, suggesting that focusing on core legal terminology balances information retention and computational efficiency. We also implemented a modified text normalization process, addressing common errors like misspellings and incorrectly merged words, which significantly improved the model's performance. These findings have important implications for automating legal document processing, aiding both professionals and the public in understanding court judgments. By leveraging advanced NLP techniques, this research contributes to enhancing transparency and accessibility in the Indonesian legal system, paving the way for more consistent and comprehensible legal decisions.**

## I. Introduction

People who are not well-versed in the law may have difficulty understanding what kinds of behaviors constitute criminal activity and the sanctions that are associated with such behavior. If it is discovered that a person has violated criminal laws, that person will be subject to sanctions and penalties that are in accordance with the applicable legal requirements. The purpose of these measures is to discourage future instances of wrongdoing via the criminal justice system.

Various entities are included in the framework of Indonesia's criminal justice system. Such entities include investigative bodies, prosecutorial institutions, judicial organs for adjudication and verdict delivery, and also penal institutions executing the decisions of these courts which are all parts of the legal system [1]. The Indonesian legal system serves as a critical tool for fighting against illegal conduct and maintaining social control. However, despite expectations of efficient and effective court operations, their actuality falls short of expectations in some cases [2]. Although legal procedure participants perceive deficiencies in the courts' competence, reliability, and ethical standards, public confidence in these judicial bodies remains significantly lower compared to other institutions [3].

In the process of adjudicating certain cases, some judges in Indonesia use jurisprudence which is previous court decisions in order to enrich the legal framework [4]. A collection of previous court decisions on similar legal issues can be used to guide their rulings on current cases [2]. It provides judges with a basis upon which they may base their decisions. The precedent that has been created by previous court judgments may be used as a means of maintaining legal certainty in situations where there is uncertainty in the law or when there are no norms that have been established [2]. The need for judicial procedures to be both transparent and consistent poses a challenge in the legal system. To address this, extensive textual documentation of previous court decisions is maintained, serving as a crucial resource for legal practitioners and decision-makers. [5].

Building on this context, crime statistics from the Criminal Investigation Agency of the Indonesian Pusiknas Bareskrim Polri indicate a year-on-year rise in crime rates between 2021 and 2023 [2]. As a result, the volume of court decision documents in Indonesia has consistently increased, with an average of approximately 100,000 documents being produced each month [4]. The proliferation of textual material presents substantial issues to information management, particularly considering the growing use of digital documents in the legal sector. Furthermore, it is difficult for judges to review and grasp each document that is relevant to a specific legal case in order to evaluate the severity of the case in comparison to other instances that have been heard in the past which makes it difficult for judges to maintain consistency in their decisions [4]. Attempting to satisfy the information requirements of this sector via the use of traditional documents and record analyses is a challenging process that calls for a significant amount of legal competence as well as a comprehensive understanding of legal language. This scenario gives rise to two primary problems such as judgments may have a tendency to be subjective as a result of the complex processing of material from various instances and judges will be subjected to increased strain and stress as a result of their enormous workloads, which would hinder their ability to provide competent decisions [5]. Moreover, people without legal expertise often struggle to understand crimes and their associated punishments, especially sentence lengths. These challenges highlight the need for an automated system. Such a system should quickly process multiple decision documents and extract key information efficiently to predict the length of sentence, bridging the gap between legal complexity and public understanding.

However, there has been a limited amount of study done on the topic of text mining of legal documents, particularly on Indonesian court rulings. As a result, this research investigates how past court rulings can help determine charges and sentencing severity using computational methods. This study aims to develop a model for predicting the length of court sentences, serving two primary purposes: first, to assist judges in maintaining decision-making consistency by providing relevant references based on predicted sentence lengths; and second, to offer guidance to those with limited legal experience by helping them understand potential sentence durations for various cases. By focusing on sentence length prediction, we aim to enhance both the consistency of judicial decisions and the public's understanding of likely legal outcomes. To conclude, this research makes several key contributions:

(1) It examines the use of Indonesian court documents to predict the duration of sentences in legal cases.
(2) The study explores a combined hybrid approach using various deep learning techniques to capture both local patterns and long-term dependencies in sentence length prediction.
(3) This work examines the impact of reduced token usage in word embeddings on the model's ability to generalize in sentence length prediction.

## II. LITERATURE REVIEW

The task of mining process entails the utilization of documents to extract insights from large datasets. In the domain of big data analytics, text mining has surfaced as a powerful instrument for capitalizing on unstructured textual data. Using text mining, it makes it easier to find new ideas and important patterns and connections that are hidden in natural language texts [6]. The challenge presented by the inherent lack of structure in textual data, which requires specialized methodologies for efficient analysis, underscores the importance of text mining techniques [2]. Text mining is an effective process that converts unstructured text into a structured format, thereby enabling the efficient analysis and identification of important patterns [7]. In general, the purpose of text mining is to extract knowledge from human-generated content and other sources of natural language data [2]. Text mining can be conducted on a wide range of document repositories, which offers limitless potential for a wide range of applications in different domains. Furthermore, text mining and document analysis have been implemented in many domain areas such as computational social science to legal [8]–[13]. However, the application of text mining in the legal domain, specifically in the Indonesian legal text research such as prediction of court rulings, has received relatively little attention as opposed to other tasks despite its extensive implementation in various fields [2].

The study of categorization and analysis of arguments in legal discourse has been an essential area of research from the beginning of the argument mining field [11]. Particularly, there has been a substantial amount of advancement in the area of Legal Artificial Intelligence (Legal AI) in recent decades [14]. Legal AI focuses on legal tasks using artificial intelligence, particularly natural language processing that uses AI to improve legal processes [15]. Many of these research has focused on extracting valuable insights from court decision documents [16]–[18]. An attempt to investigate the intersection of artificial intelligence and law by creating an argument mining model tailored to the unique difficulties posed by legal texts in European Court of Human Rights decisions dataset has been explored previously [11]. The same dataset has also been used to assess the effectiveness of GPT-like models, particularly GPT-3.5 and GPT-4, for argument mining as a classification task using prompting [6].

Multiple studies have shown the significance of court documents as a rich resource for research in the domain of legal text mining [2], [4], [5], [14]. A common objective in this area of study is to automatically categorize legal documents into different categories, providing significant benefits for legal professionals and other administrative duties. Previous study attempted to create a deep learning model that could analyze court documents and classify rhetorical structures of sentences to improve legal systems' processing time [16]. In a comparable direction, [18] presents a legal documentation classifier that utilizes conventional machine learning techniques and transformer-based neural network models, such as Legal-BERT, for the classification task. This implementation offers opportunities for legal professionals to improve the efficiency of their operations. Another example is the automated technique that uses semantic annotation to streamline the tasks of pattern recognition, information extraction, and document classification, enhancing the capacity of legal practitioners to retrieve relevant information, both within and outside the judicial system, by utilizing a supervised machine learning classifier [17].

A further advantage of analyzing court documents is that it enables classification models to acquire knowledge and predict court decisions. In a study conducted by [19], text classification techniques were utilized to assist judges and pre-trial detainees in predicting the likelihood of provisional prisoners being released by the Supreme Federal Court. This study aimed to establish a reliable association between judgment outcomes and the nature of the crime. To optimize the implementation of jurisprudence, [2] created a system for predicting lawsuit articles using previous court decisions records from Indonesian courts. Another attempt to predict court decision outcomes was investigated in [5] by categorizing drug-related criminal cases into binary classification of light and heavy classes. Similar to this, [20] developed machine learning models to predict U.S. Supreme Court case results using court documents, achieving an F1-score of 0.324 and an AUC of 0.68 for the best classification model. In a study related to Taiwan's district court, [14] employed BERT to predict the type of crime and sentence length based on court judgments. Moreover, [4] applied a CNN and attention model to Indonesian court decision documents in order to classify punishment categories. Previous research trends have used court documents to classify court decisions into predetermined classes [2], [4], [14], [19], [20]. In contrast, the current investigation conducted by [4] aims at predicting the length of the punishment rather than classifying specific categories of punishment. Thus, the experiment in this study seeks to fill this void by investigating and building upon the work in [4] to create an improved model for predicting the length of punishments. Previous research has demonstrated that various legal AI applications have been explored across diverse judicial systems. There has been significant interest in AI tasks aimed at predicting sentence lengths using court documents. Building on previous research, this study contributes to the field by focusing on predicting sentence lengths using Indonesian court documents. This approach addresses a gap in the existing literature, as there is currently limited research utilizing Indonesian legal datasets for this purpose.

## III. Materials and Methods

This section explores the methodology used from dataset description, preprocessing steps, and modelling in the experiments. Following this, several machine learning models are employed to perform regression tasks to predict the length of punishment duration. The final section specifies the evaluation techniques used in the experiment.

### *A. Dataset*

The goal of this study is to predict the length of punishments by extracting knowledge from court documents. The court documents dataset used from this study is derived from PDF files of court record documents released by Indonesian Supreme Court's website. Previous study has converted this pdf dataset into XML format, resulting in 22,650 XML documents which contain court sections commonly specified during court verdict and its corresponding verdict. Table I shows sections provided in the court documents stored in XML format as defined in [4].

TABLE I. Section Name within XML files derived from Indonesian Supreme Court's document.

| No. | Section Name |
|---|---|
| 1 | Document Opener |
| 2 | Identity of Dependant |
| 3 | Case History |
| 4 | Detention History |
| 5 | Prosecution History |
| 6 | Indictment History |
| 7 | Facts |
| 8 | Legal Facts |
| 9 | Legal Considerations |
| 10 | Verdict |
| 11 | Closing |

It is important to note that as in [4], the length of punishment is provided in Document Opener section. The length of punishments is recorded as an integer value representing the number of days sentenced. Fig. 1 illustrates the distribution of punishment durations, originally measured in days. To facilitate easier interpretation and reduce the scale, we converted these durations into years. This converted measure serves as the dependent variable in our experiments.

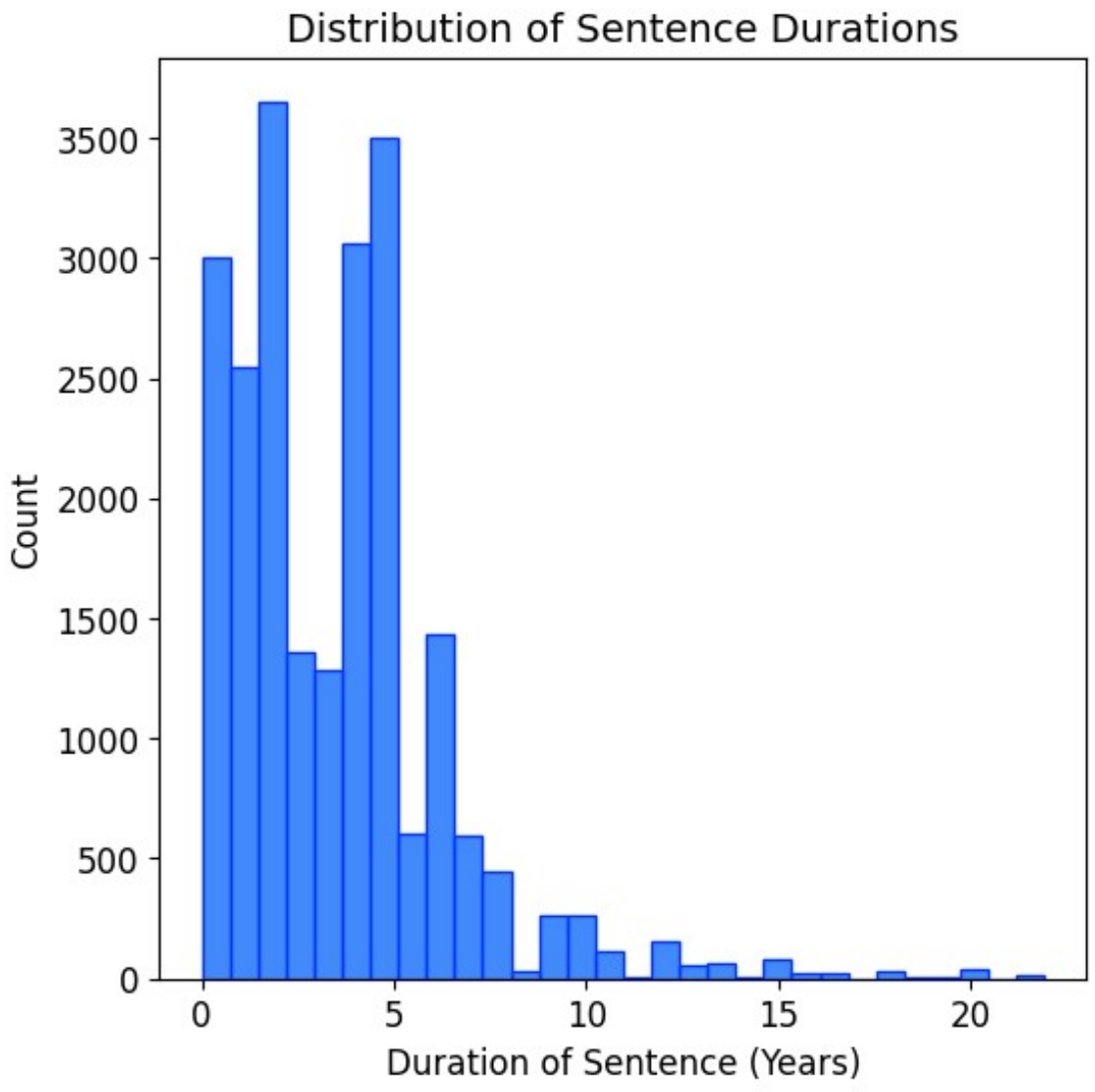

Fig. 1. The Length of Punishment Distribution.

Following this, only the Prosecution History, Facts, Legal Facts, and Legal Considerations sections are used to predict the length of punishments. This selection is based on findings from [4] which demonstrated that this combination of sections provides the highest accuracy. These sections were chosen as they represent the most relevant and substantive information pertaining to the case, encompassing the critical details that typically influence sentencing decisions.

A straightforward data cleaning process is implemented to remove documents with empty values. As a result of this cleaning process, the total number of documents in our dataset was reduced from 22,650 to 22,447 files.

### *B. Experiments*

The experiment on predicting sentence duration using court documents starts by selecting appropriate word embedding methods, specifically skipgram and CBOW, for further experiment and analysis. Skipgram predicts context words given a target word, aiming to capture word relationships. In contrast, CBOW predicts a target word based on its surrounding context, focusing on the collective meaning of nearby words [21].

To compare these embedding techniques, we employed several conventional machine learning algorithms: Random Forest, XGBoost, and Support Vector Machines (SVM). These models were chosen for their relatively low computational demands compared to deep learning approaches which is sufficient for initial experiments in determining which word embedding technique to use. Additionally, these algorithms have been widely applied in similar studies within the legal text analysis domain [22]–[24].

We assessed the models' performance using the R-squared metric, obtained through a 5-fold cross-validation process. This approach helps ensure the reliability and generalizability of our results across different subsets of the data.

To evaluate the experiments, we used the R-squared score to compare the performance between models, following the approach of several previous studies in this domain [4], [25]–[27]. R-Squared computes scores to show how well each model, using text representations of court documents as a basis to predicts the duration of punishments. Better model performance in capturing the variance of the dependent variable (punishment length) is indicated by higher R-squared scores. The R-squared scores are calculated using formula presented in Eq. (1).

$$R^2 = 1 - \frac{\sum_{i=1}^{n}(y_i - \hat{y}_i)^2}{\sum_{i=1}^{n}(y_i - \bar{y})^2} \quad (1)$$

The second experiment investigates the application of text summarization techniques to address the challenges posed by lengthy court documents as adapted from [28], [29]. This approach aims to enhance machine learning models' ability to capture relevant context from extensive legal texts when predicting sentence durations. We opted for extractive summarization as it identifies and compiles salient text segments while preserving the original wording, which is crucial for legal documents. Abstractive summarization, which generates new text, was not explored due to the sensitive nature of legal documents. These texts often contain precise language, specific terminology, and nuanced information that carries legal weight and implications. Altering the original wording could potentially change the meaning or interpretation of the document, which could lead to misunderstandings or inaccuracies in the prediction process.

Extractive summarization in this study employs TF-IDF based extractive summarization technique for the computational efficiency and effectiveness [30]. This method works by assigning importance scores to sentences based on the significance of their constituent words within the context of the entire document. The process begins by splitting the document into individual sentences. Then, a TF-IDF vectorizer is applied to convert these sentences into numerical vectors. TF-IDF weighs the importance of words by considering both their frequency within a sentence (term frequency) and their rarity across all sentences (inverse document frequency). After vectorization, the algorithm calculates a score for each sentence by summing the TF-IDF values of its words. This score represents the sentence's overall importance within the document. Finally, the sentences are ranked based on these scores, and the top-scoring sentences are selected to form the summary. This approach effectively identifies sentences that contain words that are both frequent within the sentence and distinctive across the document, likely capturing key information [30]. The resulting summary maintains the exact wording of the original text, ensuring that critical legal terminology and phrasing remain intact.

For this second experiment, we employed the same conventional models used in the first experiment: Random Forest, XGBoost, and Support Vector Machines (SVM). Each model underwent two separate training and testing phases - one using the summarized text and another using the full text. By comparing the R-squared scores between these processes, we aimed to evaluate the effectiveness of summarization in improving prediction accuracy for sentence duration. This approach allows us to assess whether focusing on key extracted information can lead to more accurate predictions compared to processing the entire document.

The third experiment delves into various deep learning architectures to predict sentence length. We explored Convolutional Neural Networks (CNN), Long Short-Term Memory (LSTM), and Bidirectional LSTM (BiLSTM) models. Our interest lies in examining whether CNNs can capture local patterns to enhance prediction performance. Additionally, we investigated LSTM and BiLSTM to assess the impact of sequential and long-term dependencies on model's performance. To further refine these architectures, we incorporated attention mechanisms to help the models focus on the most relevant textual elements.

The CNN component of our model analyzes local patterns in the input text. It begins with word embeddings representing court documents, which encode semantic relationships between words. These embeddings are then processed through multiple convolutional layers that apply filters to detect key features and phrases. We employ ReLU activation functions to introduce non-linearity, enabling the model to learn complex representations. Following the convolutional operations, a Global Max Pooling layer condenses the extracted features by selecting the most salient signals from each feature map. This results in a compact feature vector that encapsulates critical local information from the court documents. Fig. 2 illustrates this CNN architecture, showcasing the flow from input embeddings through convolutional layers to the final pooled output.

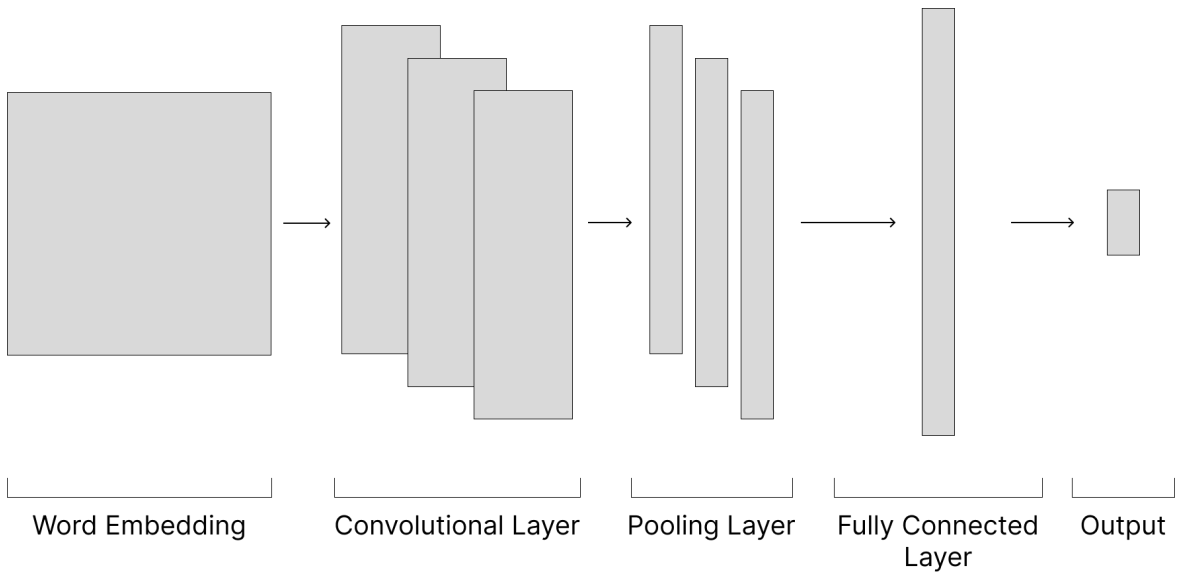

Fig. 2. Architecture of CNN

The BiLSTM branch processes the same word embeddings as the CNN but focuses on capturing long-term dependencies and contextual information. It reads the input sequence in both forward and backward directions, maintaining hidden states that reflect information from previous and future time steps. This bidirectional approach allows the model to leverage both past and future context when making predictions. To enhance its contextual understanding, we've integrated an attention mechanism into the BiLSTM architecture. This mechanism computes attention scores for each time step, enabling the model to focus on the most relevant parts of the input sequence. Fig. 3 demonstrates the BiLSTM structure with attention, highlighting how information flows bidirectionally and how attention weights are applied to the hidden states.

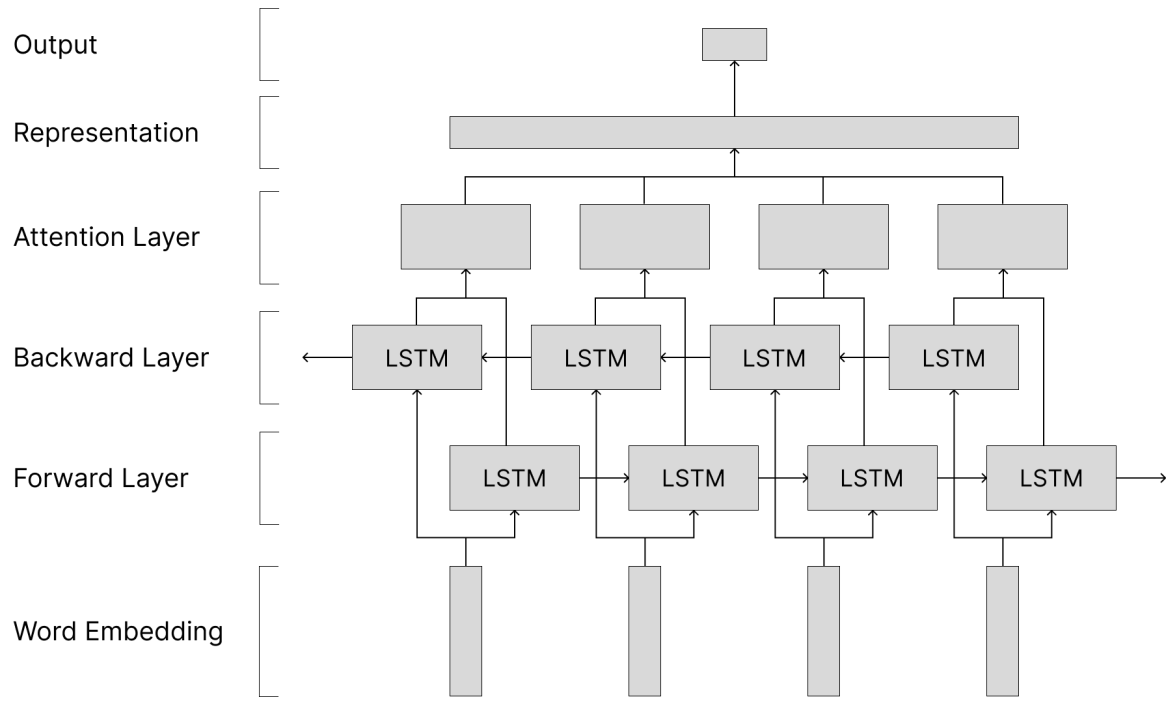

Fig. 3. Architecture of BiLSTM with Attention

The attention mechanism focuses on the most relevant parts of the input sequence during the prediction task. It operates by calculating attention scores for each time step, which are then normalized using a softmax function to create a set of weights that sum up to one. These weights determine the importance of each hidden state in the final context vector. By incorporating this context vector back into the BiLSTM output, we create a richer feature representation that emphasizes critical parts of the input sequence. Fig. 3 provides an overview of the attention mechanism layer within BiLSTM architecture, showing how attention scores are computed and applied to create the final context vector.

To potentially improve the R-squared score, we explored hybrid model which combines the strengths of both CNNs and BiLSTM models to capture more nuanced patterns in the text. This hybrid approach aims to capture both local patterns and long-term dependencies by concatenating features extracted from CNN and LSTM/BiLSTM layers before passing them to a dense layer. These methods have been used previously in similar tasks as in [31], [32]. This creates a hybrid feature representation, as illustrated in Fig. 4.

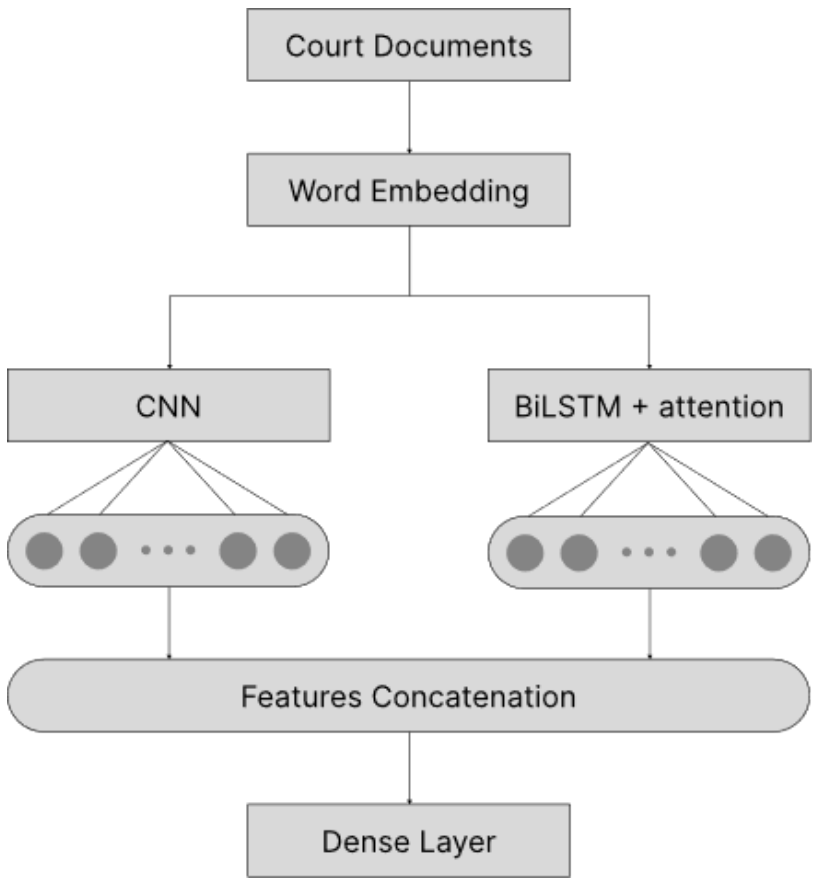

Fig. 4. The architecture of the Hybrid Model of CNN and BiLSTM + attention.

The hybrid model processes court documents in several steps. First, it converts the text into word embeddings. These embeddings then pass through two parallel branches: a CNN and a BiLSTM with attention. The CNN extracts local patterns, while the BiLSTM captures long-term dependencies. The attention mechanism helps focus on important parts of the text.

Next, the model concatenates the outputs from both branches. This creates a single feature vector that combines local and global information. Finally, a dense layer takes this vector as input. It performs the regression task, predicting the punishment duration based on the extracted features.

This structure allows the model to analyze legal documents from multiple perspectives. The CNN identifies key local features, the BiLSTM understands the overall document structure, and the attention mechanism highlights crucial text segments. The dense layer then uses this comprehensive analysis to make the final prediction.

Deep learning models typically excel with larger datasets, which aligns well with the extensive nature of court documents in this study. However, the sheer volume of data can pose challenges for further experimentation due to the high computational demands. To address this, we explored using a reduced number of tokens for generating word embeddings.

In the fourth experiment, we utilized only 30% of the global tokens for word embedding creation. The remaining tokens were transformed into an 'UNK' (unknown) label, indicating words that are not part of the core vocabulary but may still appear in the documents. This approach helps manage vocabulary size while still capturing the most frequent and potentially most informative words.

To implement this, we first calculated the frequency of each token across all documents. We then retained the top 30% most frequent tokens and converted the remaining 70% into the UNK label. This method allows us to focus on the most common and potentially most significant terms while reducing computational complexity.

To further investigate the impact of vocabulary size on model performance, the experiment is expanded to include different cut-off percentages for token usage. A range of cut-offs were selected: 10%, 30%, 50%, 70%, and 100%. It's worth noting that the 30% and 100% scenarios were already explored in third and fourth experiments, providing a basis for comparison.

This approach allows us to analyze how varying levels of vocabulary reduction affect the R-squared score of our models. By comparing the performance across these different cut-offs, we can identify the optimal balance between vocabulary size and prediction accuracy, potentially revealing insights into the trade-off between computational efficiency and model performance in legal text analysis tasks.

The fifth experiment delves into text normalization. Previous research in this domain has explored a two-step normalization process to enhance the quality of input data for predicting sentence length. This process addresses common issues in legal documents, such as typos and incorrectly combined words.

The first step focuses on typo correction. As noted in [4], typos can originate from either the document's original creator or during the PDF into XML conversion process. The correction method works as follows:

Tokens with a frequency of 10 or higher are considered correct and stored in a dictionary D. For tokens appearing less than 10 times in a document, numeric tokens are converted to UNK (unknown). Non-numeric infrequent tokens are compared to entries in dictionary D using minimum edit distance. If this distance is less than 3, the token is replaced with the closest match from D; otherwise, it's converted to UNK. This process is illustrated in the step 1 Typo Correction section of Fig. 5.

The second normalization step involves space splitting, which builds upon the output of the typo correction phase. This process aims to address instances where words have been incorrectly combined. It works by systematically adding spaces to each token, starting from the first letter and moving rightward. At each position, the algorithm checks if inserting a space would create a new pair of tokens that exist in dictionary D. This method helps identify and correct cases where two valid words have been merged into a single incorrect token. This process is illustrated in the step 2 Space Splitting section of Fig. 5.

These sequential normalization steps are applied to create a refined dataset before inputting it into the machine learning models for sentence length prediction. This preprocessing approach aims to improve data quality and, consequently, the accuracy of the prediction models by addressing common textual issues in legal documents.

These normalization steps are adopted from [4], with an additional step inserted to handle infrequent numeric tokens. Specifically, infrequent numeric tokens are converted to UNK (unknown). This modification is based on the assumption that infrequent numbers might not carry significant weight in the prediction model. By converting these to UNK, we potentially reduce noise in the data while retaining the most relevant numerical information. This adaptation tailors the normalization process to the specific needs of legal text analysis, where certain numeric details may be less crucial for punishment duration prediction than the overall textual content and frequently occurring numbers.

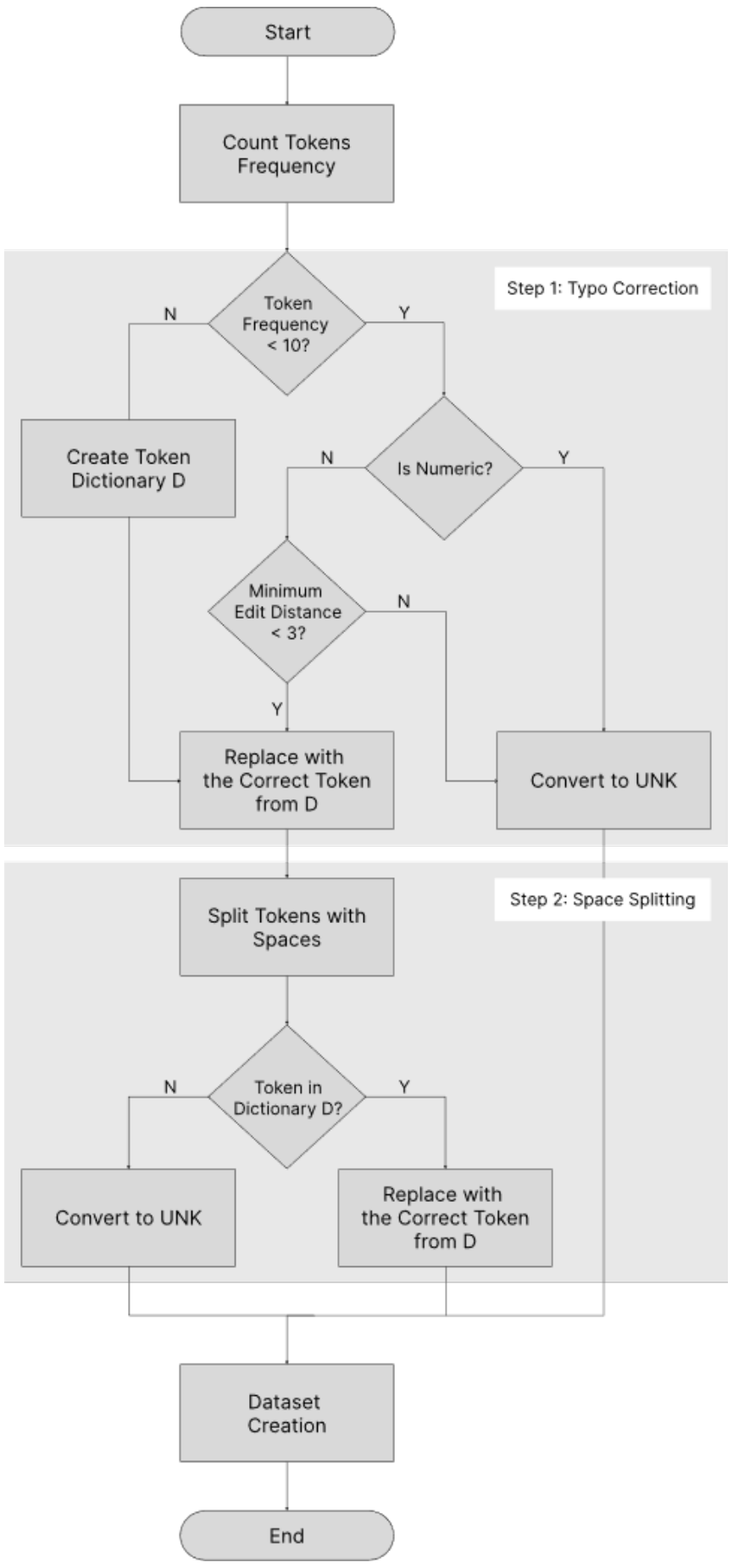

Fig. 5. Two sequential steps of text normalization: Typo Correction and Space Splitting.

Table II provides a comprehensive overview of the different model configurations and experiments conducted in this phase of the study, highlighting the various architectures and their combinations explored to optimize punishment duration prediction in legal documents.

TABLE II. SUMMARY OF EXPERIMENTAL SCENARIOS AND METHODS FOR PREDICTING PUNISHMENT DURATIONS IN INDONESIAN COURT RULINGS

| No | Experiment | Method | Evaluation Metric |
|---|---|---|---|
| 1. | Word Embedding Comparison (Skipgram vs CBOW) | Random Forest<br>XGBoost<br>SVM | R-squared score |
| 2. | Full Text vs Summary | | |
| 3. | Deep Learning Models | CNN<br>CNN + attention<br>LSTM<br>LSTM + attention<br>BiLSTM<br>BiLSTM + attention<br>Hybrid (CNN, BiLSTM)<br>Hybrid (CNN, BiLSTM + attention) | |
| 4. | Token Usage Optimization (10%, 30%, 50%, 70%, 100%) | | |
| 5. | Text Normalization | Hybrid (CNN, BiLSTM + attention) | |

This study investigates five key aspects of predicting punishment durations in Indonesian court rulings. We compare: (1) Skipgram and CBOW word embedding techniques, (2) full text versus summarized text analysis, (3) various deep learning architectures including CNN, LSTM, BiLSTM, and hybrid models with attention, (4) the impact of using different percentages of the most frequent tokens in our hybrid model, and (5) the effect of text normalization through typo correction and space splitting. All experiments use the R-squared score for consistent evaluation, allowing us to systematically assess the most effective approach for this legal text analysis task.

## IV. RESULT AND DISCUSSION

For the first experiment, we compared the performance of CBOW and skip-gram word embedding techniques in predicting sentence length using court documents. The results show that skip-gram consistently outperforms CBOW across all three models (Random Forest, XGBoost, and SVM) in predicting sentence length from court documents as seen Fig. 6. Despite improving the R-squared score by just under 0.1, Skip-gram achieves higher R-squared scores in each model.

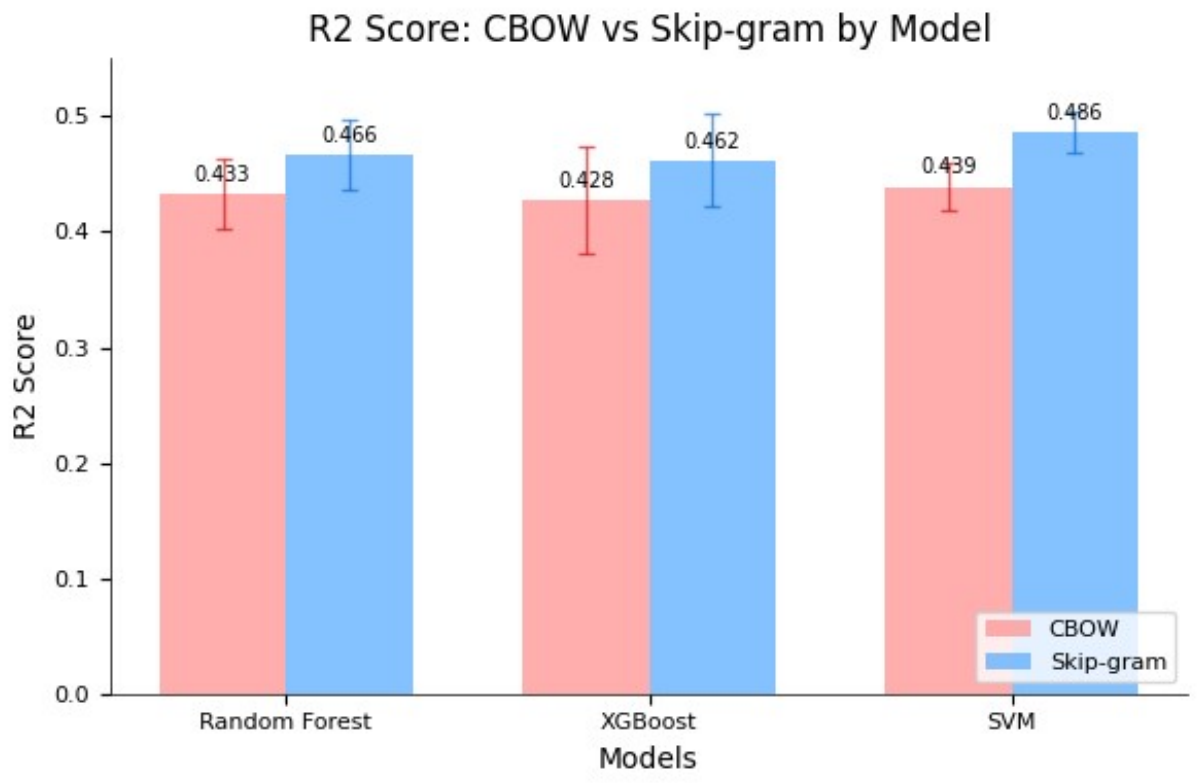

Fig. 6. Comparison of R2 Scores between CBOW and Skip-gram Word Embeddings across Different Machine Learning Models

Skip-gram's superior performance likely stems from its ability to capture more nuanced word relationships. Unlike CBOW, which predicts a target word from its context, skip-gram predicts surrounding context words given a target word. This approach may be more effective in legal texts where specific terms can have significant impact on sentence outcomes.

For the second experiment, we investigated the impact of document summarization on predicting sentence length using the same three models: Random Forest, XGBoost, and SVM. We compared the R-squared scores of models trained on full text documents versus summarized documents, all using skip-gram word embeddings. The results of this comparison are presented in Table III.

TABLE III. COMPARISON OF R-SQUARED SCORES AND STANDARD DEVIATIONS FOR FULL TEXT VS. SUMMARIZED DOCUMENTS USING SKIP-GRAM WORD EMBEDDINGS

| Model | Text Type | R-squared score | Standard Deviation |
|---|---|---|---|
| Random Forest | Full text | **0.4655** | 0.0303 |
| | Summarize | 0.3454 | **0.0088** |
| XGBoost | Full text | **0.4619** | 0.0401 |
| | Summarize | 0.3170 | **0.0291** |
| SVM | full text | **0.4860** | **0.0179** |
| | summarize | 0.3679 | 0.0290 |

The findings consistently show that models trained on full text outperform those trained on summarized text in terms of R-squared scores. Random Forest achieved 0.4655 (full text) vs 0.3454 (summarized), XGBoost 0.4619 vs 0.3170, and SVM 0.4860 vs 0.3679. This suggests that full text documents contain valuable information for sentence length prediction that is lost in summarization.

Interestingly, for Random Forest and XGBoost, models trained on summarized text showed lower standard deviations in their R-squared scores. This could indicate that summarization provides a more consistent, albeit less informative, representation of the documents. Although SVM has the opposite trend as it maintains lower standard deviation with full text.

The superior performance of full text models, despite higher computational costs, implies that the nuances and details in legal documents are crucial for accurate sentence length prediction. The summarization process, while potentially useful for other tasks, appears to omit important signals for this specific prediction task. The consistent pattern across different model types strengthens the conclusion that full text is preferable for this task, regardless of the chosen algorithm. This might suggest that legal language requires comprehensive context for accurate interpretation in automated systems.

For the third experiment, we explored various deep learning models to predict sentence length from court documents. Table IV shows the R-squared scores for these models.

TABLE IV. R-SQUARED SCORES OF DEEP LEARNING MODELS FOR SENTENCE LENGTH PREDICTION USING FULL VOCABULARY

| Model | R-squared score |
|---|---|
| LSTM | 0.5592 |
| LSTM + attention | 0.5513 |
| CNN | 0.5331 |
| CNN + attention | 0.5214 |
| BiLSTM | 0.5557 |
| BiLSTM + attention | **0.5754** |
| Hybrid (CNN, BiLSTM) | 0.5403 |
| Hybrid (CNN, BiLSTM+attention) | 0.5167 |

The results indicate that LSTM-based models (LSTM, LSTM+attention, BiLSTM, BiLSTM+attention) generally outperform CNN-based models which is reasonable. The reason is that the sequential nature of legal text is better captured by LSTM architectures, which are designed to handle long-term dependencies in sequential data. Legal documents often contain complex, interconnected ideas that span multiple sentences, which LSTM based architectures may be better equipped to process.

Interestingly, attention mechanisms only improved performance for BiLSTM, with BiLSTM+attention achieving the highest R-squared score of 0.5754. This implies that bidirectional processing combined with attention allows the model to focus on the most relevant parts of the document for sentence length prediction, enhancing its performance.

The fourth experiment uses only the top 30% of tokens by frequency and we observed a shift in performance such as all models maintained or improved their R-squared scores; despite using fewer tokens and the hybrid model (CNN, BiLSTM+attention) achieved the highest R-squared score of 0.5893 as can be observed in Table V.

TABLE V. R-SQUARED SCORES OF DEEP LEARNING MODELS FOR SENTENCE LENGTH PREDICTION USING TOP 30% FREQUENT TOKENS

| Model | R-squared Score |
|---|---|
| LSTM | 0.5520 |
| LSTM + attention | 0.5588 |
| CNN | 0.5731 |
| CNN + attention | 0.5710 |
| BiLSTM | 0.5543 |
| BiLSTM + attention | 0.5800 |
| Hybrid (CNN, BiLSTM+attention) | **0.5893** |

These results suggest that using a reduced token set may actually benefit the models, possibly by reducing noise and focusing on the most informative words. The improved performance of hybrid models with fewer tokens implies that combining the local feature detection of CNNs with the sequential processing of BiLSTMs can effectively capture both short-range and long-range dependencies in legal text, even with a limited vocabulary.

This finding has important implications for practical applications. First, it demonstrates that effective models can be built with reduced computational resources. Second, it suggests that legal text analysis might benefit from focusing on key, frequently used terms rather than processing entire vocabularies. Lastly, the success of hybrid models indicates that combining different neural network architectures can lead to more robust and accurate predictions in legal document analysis.

In the follow up experiment, we investigated the impact of token usage percentage on model performance using the Hybrid (CNN, BiLSTM + attention) model, which previously showed the best results. Fig. 7 illustrates the relationship between the percentage of tokens used and the corresponding R-squared scores.

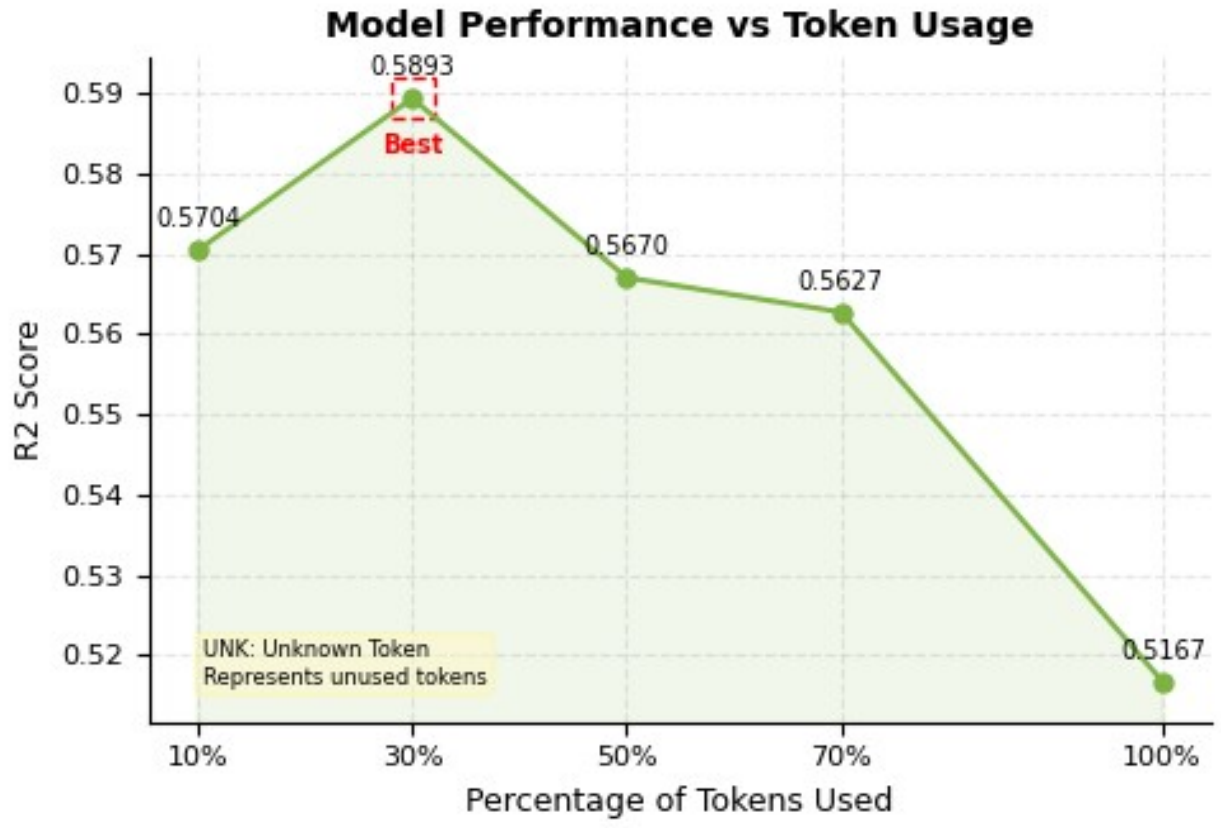

Fig. 7. Impact of Token Usage Percentage on Model Performance.

The results show an interesting pattern, with 30% token usage yielding the highest R-squared score of 0.5893, followed closely by 10% usage at 0.5704. Performance gradually declined as token usage increased beyond 30%, with 100% token usage resulting in the lowest R-squared score of 0.5167. These findings suggest that focusing on the most frequent tokens in legal documents can actually improve prediction accuracy while significantly reducing computational resources.

This counterintuitive result might be explained by noise reduction, as less frequent words may introduce noise rather than valuable information for sentence length prediction. Additionally, the most frequent tokens likely represent core legal concepts and terminology crucial for determining sentence length. Using fewer tokens may also help the model avoid overfitting to specific document peculiarities, leading to improved generalization.

The superior performance at 30% token usage indicates an optimal balance between information retention and noise reduction. This has important implications for practical applications in legal text analysis, potentially enabling more efficient and scalable processing of larger document sets or real-time analysis with limited resources. Furthermore, focusing on a smaller set of key tokens might lead to more interpretable models, which is crucial in the legal domain.

These results highlight the potential for developing more efficient and effective models for legal document analysis by carefully selecting the most informative tokens rather than using the entire vocabulary. This approach could significantly impact how we process and analyze legal documents, offering a path to improved performance with reduced computational demands.

The experiment comparing the hybrid (CNN, BiLSTM + attention) architecture with and without text normalization yielded interesting results. The normalized version achieved an R-squared score of 0.5551, significantly outperforming the non-normalized version's score of 0.5167 as seen in Table VI. This improvement highlights the crucial role of preprocessing in enhancing prediction accuracy for sentence length in legal documents.

TABLE VI. IMPACT OF TEXT NORMALIZATION ON HYBRID MODEL PERFORMANCE

| Text Preprocessing | Model Architecture | R-squared Score |
|---|---|---|
| Full text | Hybrid (CNN, BiLSTM+attention) | 0.5167 |
| Full text + Normalization | Hybrid (CNN, BiLSTM+attention) | **0.5551** |

The normalization process, which includes typo correction and space splitting, appears to address several key issues in legal text data. The typo correction step, by replacing infrequent tokens with either their closest match in a frequency-based dictionary or converting them to UNK, likely reduces noise in the data. This process may help the model focus on the most reliable and relevant information, filtering out potential errors introduced during document creation or conversion.

The space splitting step tackles incorrectly combined words, a common issue in digitized legal documents stemming from both human errors and technical glitches in the PDF to XML conversion process [4]. By systematically separating merged words, this normalization technique addresses unintentional errors that could significantly alter the meaning of legal text. This correction enhances the model's ability to accurately interpret the document's content, leading to improved sentence length predictions. Essentially, the normalization process acts as a crucial filter, mitigating the impact of human and technical errors on the model's performance, thereby improving the overall accuracy of legal text analysis.

The substantial improvement in R-squared score suggests that these preprocessing steps effectively clean and standardize the input data, allowing the model to learn more meaningful patterns. This is particularly important in legal texts, where precision in language is crucial, and small errors or inconsistencies can significantly impact interpretation.

While the preprocessing steps are computationally demanding, the results indicate that they provide valuable benefits for prediction accuracy. This trade-off between computational cost and improved performance is an important consideration in developing practical applications for legal text analysis. The significant improvement in results may justify the additional computational resources required, especially in contexts where high accuracy is critical.

These findings underscore the importance of thorough text preprocessing in legal document analysis and suggest that investing in robust data cleaning and normalization techniques can lead to substantial improvements in model performance.

## V. CONCLUSION

This study explored various approaches to predict court punishment duration using Indonesian legal documents, aiming to develop a system that assists judges in maintaining decision consistency and guides those with limited legal experience. Our experiments yielded several significant findings. The hybrid architecture combining CNN and BiLSTM with attention mechanism demonstrated superior performance, achieving the highest R-squared score of 0.5893. This model's success likely stems from its ability to capture both local patterns through CNN and long-term dependencies via BiLSTM, providing a comprehensive analysis of legal text crucial for sentence prediction. Interestingly, while document summarization did not improve accuracy, using only the top 30% most frequent tokens improved prediction performance. This suggests that focusing on core legal terminology, rather than full document summarization, effectively balances information retention and computational efficiency. The method of selective vocabulary reduction proved more effective than summarization in maintaining essential context for prediction. Finally, our text normalization process, addressing issues like typos and incorrectly merged words, significantly boosted prediction accuracy, highlighting the importance of preprocessing in legal text analysis. These results underscore the potential of advanced NLP techniques in automating and improving legal document processing, particularly in the context of Indonesian court systems. Such advancements could not only aid legal professionals but also enable the public to better understand and scrutinize court judgments, furthering the study's broader goals of enhancing transparency and accessibility in the legal system.

Future research should focus on optimizing and expanding normalization techniques, exploring new preprocessing methods to enhance legal text data quality. Another significant avenue for advancement lies in leveraging more advanced language models such as BERT or other LLMs, which were beyond the scope of this study due to computational limitations. Fine-tuning these models on Indonesian legal corpora could potentially capture more nuanced legal language patterns and improve model's performance. Additionally, investigating the interpretability of these models remains crucial in the legal domain, where understanding the reasoning behind predictions is paramount. These directions aim to further enhance the accuracy and applicability of automated sentence length prediction in Indonesian legal contexts.

## Conflict of Interest

The authors declare no conflict of interest.

## Author Contributions

MAI, ATH, MSA conducted the research. MAI and ATH analyzed the data and implemented the experiment. MSA validated the result. MAI and ATH wrote the original draft. MSA supervised the overall process. All authors had approved the final version.